\documentclass{article}

    \PassOptionsToPackage{round}{natbib}

\usepackage{xcolor}
\usepackage[hidelinks]{hyperref}
\definecolor{linkcolor}{RGB}{0, 0, 128}
\hypersetup{
     colorlinks   = true,
     citecolor    = linkcolor,
     linkcolor    = linkcolor,
     urlcolor     = linkcolor,
}

     \usepackage[preprint]{neurips_2024}

\usepackage[utf8]{inputenc} 
\usepackage[T1]{fontenc}    
\usepackage{hyperref}       
\usepackage{url}            
\usepackage{booktabs}       
\usepackage{amsfonts}       
\usepackage{nicefrac}       
\usepackage{microtype}      
\usepackage{xcolor}         
\usepackage{multirow}
\usepackage{amsmath}
\usepackage{graphicx}
\usepackage{subcaption}
\usepackage{tcolorbox}
\usepackage{adjustbox}
\usepackage{todonotes}
\usepackage{mathtools}
\usepackage{bbm}
\usepackage{bm}

\title{Uncertainty-aware Step-wise Verification \\ with Generative Reward Models}

\author{%
\bf Zihuiwen Ye$^{1}$\thanks{Correspondence to \texttt{zihuiwen.ye@cs.ox.ac.uk}.}\hspace{0.2em}, \hspace{0.1em} Luckeciano Carvalho Melo$^{1}$\thanks{Equal contribution.}\hspace{0.2em}, \hspace{0.1em} Younesse Kaddar$^{1}$$^{\dagger}$, \\[0.3cm]
\bf Phil Blunsom$^{2}$, Sam Staton$^{1}$, Yarin Gal$^{1}$ \\[0.3cm]
$^{1}$University of Oxford, $^{2}$Cohere \\
}

\begin{document}

\maketitle

\begin{abstract}
Complex multi-step reasoning tasks, such as solving mathematical problems, remain challenging for large language models (LLMs). While outcome supervision is commonly used, process supervision via process reward models (PRMs) provides intermediate rewards to verify step-wise correctness in solution traces. However, as proxies for human judgement, PRMs suffer from reliability issues, including susceptibility to reward hacking.
In this work, we propose leveraging uncertainty quantification (UQ) to enhance the reliability of step-wise verification with generative reward models for mathematical reasoning tasks. We introduce CoT Entropy, a novel UQ method that outperforms existing approaches in quantifying a PRM’s uncertainty in step-wise verification. Our results demonstrate that incorporating uncertainty estimates improves the robustness of judge-LM PRMs, leading to more reliable verification.
\end{abstract}

\section{Introduction}

Large Language Models (LLMs) have shown impressive reasoning abilities on complex tasks by producing step-by-step solutions in a chain-of-thought (CoT) format \citep{wei2022chain}. A common approach to further improve these capabilities is to fine-tune LLMs using reinforcement learning (RL) on generated CoT outputs \citep{pang2024iterative}. However, applying RL to reasoning tasks introduces a new challenge: credit assignment. In \emph{outcome-only} reward settings, where rewards are assigned only based on the final answer, the model may receive a high reward even when its intermediate steps are flawed. This results in false positives \citep{zhang2024genrm}, where correct answers are reached by spurious reasoning, and erodes the trust in the reasoning process \citep{zhang2025lessons}. 

To address this issue, recent work has proposed \emph{process supervision} with Process Reward Models (PRMs) that provide feedback at each reasoning step, offering more fine-grained reward signals than simply judging a final answer \citep{lightman2023let, wang2024mathshepherdverifyreinforcellms, uesato2022solving}. Benchmarks like \textsc{ProcessBench} \citep{zheng2024processbench} aim to evaluate PRMs' ability to discern step-wise correctness in multi-step solution traces, and while there has been promising progress, two major problems still limit the effectiveness of PRMs. First, obtaining high-quality step-by-step annotations is challenging: current efforts relying on human annotation \citep{lightman2023let}, Monte Carlo sampling \citep{wang2024mathshepherdverifyreinforcellms}, or LLM-as-a-judge \citep{zheng2023judging} are either costly or noisy. Second, since PRMs are ultimately a learned proxy, they are susceptible to over-optimization and reward hacking: a policy may learn to exploit imperfections in the reward model to get high reward without truly improving its reasoning \citep{deepseekai2025deepseekr1}. 

In this paper, we argue that \emph{uncertainty quantification} (UQ) offers a principled way to improve the reliability of these generative reward models. By detecting when the verifier itself is uncertain at a given step, we can better guard against erroneous feedback and improve interpretability of the reward signals. While recent work has explored UQ for LLMs on simpler tasks such as question answering \citep{shorinwa2024surveyuncertainty}, to our knowledge, little is known about how to apply UQ in complex, multi-step reward modeling, where errors may arise at any intermediate step. We take a first step toward closing this gap by proposing an uncertainty-aware verification framework on mathematical reasoning tasks. Our contributions are summarized as follows:

\begin{itemize}
\item We propose a novel UQ method for step-wise verification in generative reward models, leveraging CoT-based entropy to capture uncertainty in the verifier's reasoning.
    \item We show that incorporating uncertainty estimates leads to more robust verification performance for reward models as judges on multi-step reasoning tasks. 
    \item We compare our method with existing UQ methods and show better performance as evaluated by metrics such as AUC, rejection-F1, and selective performance score.
    \item We offer insights into the main sources of PRM errors, in the context of verifying multi-step math solutions, by decomposing predictive uncertainty into knowledge uncertainty and aleatoric uncertainty.
\end{itemize}

\section{Problem Statement}
\subsection{Background on LLM Reasoning and Reward Verification}
We follow the standard setup in LLM-based reasoning. Given a LLM representing a policy $\pi$, referred to as `generator', and an input question $Q$, $\pi$ generates an output sequence $\textbf{s} \coloneq (s_1, s_2, \cdots, s_K)$ by autoregressively sampling tokens from $\pi(- \mid Q)$. For every $t \in \{1, …, K\}$, let $\textbf{s}_{<t}$ (resp. $\textbf{s}_{\leq t}$) be the subsequence $(s_1, …, s_{t-1})$ (resp. $(s_1, …, s_t)$), and let $\textbf{s}_{<1} = \textbf{s}_{\leq0}$ be the empty sequence. The conditional probability of generating an output sequence $\textbf{s}$ is given by: 
\begin{equation}
    \pi(\textbf{s} \mid Q) = \prod_{t=1}^K \pi(s_t \mid Q, \textbf{s}_{<t}).
    \label{eq:reasoning_policy}
\end{equation}

In the context of reasoning tasks, the sequence $\textbf{s} \coloneq (s_1, s_2, \cdots, s_K)$ represents a reasoning path, where $s_t$ represents the step at time $t$ (for $1 \le t \le K$) and $K$ is the total number of steps in $\textbf{s}$. The final answer derived from this reasoning path $\textbf{s}$ is denoted by $a_{\textbf{s}}$. Typically, each question $Q$ has a ground truth final correct answer $a_Q$. The model is considered to have correctly answered $Q$ if its final conclusion $a_{\textbf{s}}$ matches $a_Q$.

Given a reasoning path $\textbf{s}$ generated by a policy $\pi(- \mid Q)$, verifiers or reward models (RMs) are commonly used to evaluate the quality of $\textbf{s}$ by assigning a reward score $r(Q, \textbf{s}) \in \mathbb{R}$. Once trained, RMs can be deployed to either update $\pi$ in the RL training process, or to guide inference-time search. Specifically, given a question $Q$ and a reasoning path $\textbf{s}$, a verifier parameterized by $\theta$ produces a reward score $r_\theta(Q, \textbf{s})$ that estimates the correctness of $\textbf{s}$. This reward can depend solely on the final derived answer $a_{\textbf{s}}$ (as in outcome reward models, ORMs), or on the entire reasoning trace $\textbf{s}$ (as in process reward models, PRMs).  In this work, we focus on PRMs. PRMs are step-wise verifiers that assess the correctness of each step $s_t$ in $\textbf{s}$ \citep{lightman2023let, wang2024mathshepherdverifyreinforcellms, uesato2022solving}, allowing for more fine-grained supervision over the reasoning process than ORMs. They mitigate false positives, where a path leads to the correct solution $a_Q$ but through flawed reasoning \citep{zhang2024genrm, zhang2025lessons}. However, one key challenge with developing PRMs lies in the complex and costly process of annotating step-wise labels, leading to less robust training.

\subsection{Step-wise Verification with Generative PRM}
In this work, we focus on generative PRMs \citep{lightman2023let}, which are trained through a next-token prediction objective that maximizes the likelihood of special tokens indicating step correctness. Formally, we define a random variable with two classes $E_t \in \{0, 1\}$ at each step $s_t$ to indicate whether the step contains an error, where $E_t = 0$ means the step is error-free (correct), and $E_t = 1$ means the step contains an error. 
For a generative PRM parameterized by $\theta$, the model outputs a predictive distribution $p_\theta\bigl(E_t \mid Q, \textbf{s}_{\leq t}\bigr)$ at each time step $t$, reflecting its belief about whether step $s_t$ is correct or incorrect. From this distribution, we define the step-wise reward \(r_t\) as the probability that \(E_t = 0\):
\begin{equation}
    r_t \coloneq r_\theta(Q, s_t) = p_\theta(E_t = 0 \mid Q, \textbf{s}_{\leq t}) 
    = p_\theta \bigl([\texttt{no\_error}] \mid Q, \textbf{s}_{\leq t}\bigr),
    \label{eq:stepwise_noerror}
\end{equation}
where $[\texttt{no\_error}]$ is a special token denoting that the step contains no error. By extension, the solution-level reward is given by the product of these step-wise probabilities:
\begin{equation}
    r_\theta (Q, \textbf{s}) = \prod^K_{t=1} r_\theta(Q, s_t) = \prod^K_{t=1} p_\theta(E_t=0 \mid Q, s_{\leq t}). 
    \label{eq:prm_score}
\end{equation}
Finally, to obtain a discrete prediction for each step $t \in \{1, …, K\}$, we select the most likely class from the predictive distribution:
\begin{equation}
 \hat{e}_t = \mathrm{argmax}_{e \in \{0, 1\}} p_\theta\bigl( E_t = e\mid Q, \textbf{s}_{\leq t}\bigr),
\end{equation}
producing a sequence prediction $\hat{\mathbf{e}} = (\hat{e}_1, \dots, \hat{e}_K) \in \{0,1\}^K$ for the reasoning trace $\mathbf{s} = (s_1, \dots, s_K)$. This is then compared to the ground truth sequence labels $\textbf{y} = (y_1, y_2, \dots,y_K) \in \{0, 1\}^K$. We define the step-wise verification accuracy with an indicator function $ \mathbbm{1} (\hat{e}_t= y_t)$, which is 1 if the predicted value equals the ground truth and 0 otherwise. In practice, datasets are annotated in such a way that the reasoning trace terminates at the first error (i.e., at the first index $t$ such that $s_t$ has a ground truth label of 1). Consequently, if we assume a constant error rate at each step, the index of the first positive ground-truth label follows a geometric distribution.

\subsection{Uncertainty Quantification with Judge-LM}

We now introduce the task of uncertainty quantification for step-wise verification. In the following, we will use an LLM with parameterized policy $p_\theta$ ($\theta$ denoting its weights) as a generative PRM, referred to as a `judge-LM'. Following the LLM-as-a-judge setup \citep{zheng2023judging}, we use a prompt $\mathbf{I}$ that asks for the model's verification judgment at each step $s_t$, such as ``\texttt{Does this step contain an error?}". 
Let $\tau(e)$ denote the token corresponding to the label $e\in\{0,1\}$, 
for instance, $\tau(0) = \texttt{no}$ and $\tau(1) = \texttt{yes}$.
We want to quantify the uncertainty in the judge-LM's decision, and employ methods to detect when the LM is likely to make a mistake. To achieve this, we derive \textit{uncertainty estimates} $u_t \coloneq g(p_\theta(E_t = e \mid Q, \textbf{s}_{\leq t}, \mathbf{I}))$ from the predictive distribution $p_\theta$, where $g(\cdot)$ is an uncertainty-mapping function. 
Specifically, $p_\theta$ over the binary variable $E_t$ is obtained by:
\begin{equation}
    p_\theta\bigl(E_t = e \mid Q, \textbf{s}_{\leq t}, \mathbf{I}\bigr)
    \;=\;
    \frac{p_\theta(\tau(e) \mid Q, \textbf{s}_{\leq t}, \mathbf{I}))}
         {\displaystyle \sum_{e' \in \{0,1\}} p_\theta(\tau(e') \mid Q, \textbf{s}_{\leq t}, \mathbf{I}))},
    \quad e \in \{0,1\},
    \label{eq:pred_distr}
\end{equation}
where the numerator is the token probability assigned by the LM to the token $\tau(e)$. Thus, we obtain a well-defined normalized probability distribution over the possible label outcomes $e\in\{0,1\}$.

In the following sections, we will investigate how well different uncertainty estimates $u_t$ align with the model's actual correctness at each step (given ground-truth labels in $\{0,1\}^K$), paving the way for uncertainty-aware verification.

\section{Chain-of-Thought Entropy}


Since a better-calibrated predicted probability distribution $p_\theta$ can yield more reliable confidence estimates $u$ \citep{guo2017calibration, cattelan2023fix}, we first improve the estimation given by $p_\theta$ with chain-of-thought (CoT) prompting \citep{wei2022chain}, which is shown to improve performance in a wide range of tasks. Since verification involves nuanced reasoning, judge-LM verifiers naturally benefit from CoT, which can generate intermediate rationales or critiques to help identify subtle errors that might otherwise be missed by direct verifiers, before deciding on the correctness of a solution \citep{ye2024improvingrewardmodelssynthetic, zhang2024genrm}. In a step-wise verification setting, CoT ensures that reasoning remains focused on assessing the local logical consistency of each atomic step $s_t$ with a binary decision, while improving interpretability by providing step-wise feedback on the mistakes. Specifically, we design a prompt $\textbf{I}_{\text{CoT}}$ (shown in App.~\ref{app:prompt}) that asks the model to first output a rationale $c$, and then output an evaluation $e$ of whether the step contains an error. 

Our method is inspired by semantic entropy (SE) \citep{Farquhar2024se}, which addresses the fact that one idea can be syntactically expressed in multiple ways, by computing uncertainty at the level of semantic outcomes rather than specific sequences of words. More precisely, this is achieved by sampling and grouping semantically equivalent rationale sequences into clusters, aggregating their probabilities, and computing entropy over these semantic clusters.

We now introduce our CoT Entropy method, aiming to estimate the uncertainty of the judge-LM over its step-wise verification judgments, conditioned on diverse reasoning paths. Intuitively, our method samples several reasoning paths leading to binary judgments for each step (whether this step is deemed incorrect or not), and then marginalizes all these reasoning paths to compute an estimate of the entropy over the verification judgment.

To simplify the notations, we define $\mathbf{x}_{\leq t} \coloneq (Q, \textbf{s}_{\leq t}, \mathbf{I}_{\text{CoT}})$, which represents the context up to step $t$, including the input question $Q$ and the CoT prompt $ \mathbf{I}_{\text{CoT}} $. Therefore, the entropy over the predictive distribution (Eq.~\ref{eq:pred_distr}) can be expressed as:
\begin{equation}
    \mathrm{H}(E_t \mid Q, \textbf{s}_{\leq t}, \mathbf{I}_{\text{CoT}}) = \mathrm{H}(E_t \mid \mathbf{x}_{\leq t}) = - \textstyle\sum \limits_{e} p_\theta(e \mid \mathbf{x}_{\leq t}) \log p_\theta(e \mid \mathbf{x}_{\leq t}), \quad e \in \{0,1\}.
\end{equation}
CoT Entropy is an approximation of this entropy by marginalizing over some rationales $c$: 
\begin{equation}
\begin{aligned}
    \mathrm{CoTE}(\mathbf{x}_{\leq t}) &= - \textstyle\sum \limits_{e} \Bigl(\Bigl[ \textstyle\sum \limits_{c} p_\theta(c, e \mid \mathbf{x}_{\leq t}) \Bigr] 
    \log \Bigl[ \textstyle\sum \limits_{c} p_\theta(c, e \mid \mathbf{x}_{\leq t}) \Bigr]\Bigr) \\
    &= - \textstyle\sum \limits_{e} \Bigl(\Bigl[ \textstyle\sum \limits_{c} p_\theta(e \mid \mathbf{x}_{\leq t}, c) p_\theta(c \mid \mathbf{x}_{\leq t}) \Bigr] 
    \log \Bigl[\textstyle\sum \limits_{c} p_\theta(e \mid \mathbf{x}_{\leq t}, c) p_\theta(c \mid \mathbf{x}_{\leq t}) \Bigr]\Bigr),
\end{aligned}
\label{eq:cot_entropy}
\end{equation}
where a reasoning path $c$ is sampled given a context $\textbf{x}_{\leq t}$, followed by a verification decision $e$ sampled conditioned on $c$: 
\begin{equation}
c \sim p_\theta(-\mid \mathbf{x_{\leq t}}),\quad e \sim p_\theta(- \mid c).
\end{equation}
In practice, we follow three steps: 
\begin{enumerate}
    \item \textit{Generation:} Given a judge-LM and a context $\textbf{x}_{\leq t}$, sample reasoning sequences $c$ that critique the current step $s_t$, followed by a decision $e$ (of whether $s_t$ is deemed incorrect) after the CoT.
    \item \textit{Clustering:} Group the output sequences into two clusters based on whether the decision $e$ equals $1$ (deemed incorrect) or $0$ (deemed correct).
    \item \textit{Entropy estimation:} Normalize token probabilities $\tau(e)$ to obtain class probabilities using Eq.~\ref{eq:pred_distr} for each sequence in both clusters. Then, sum the class probabilities of sequences leading to the same decision $e$ and compute the resulting entropy.
\end{enumerate}

We highlight that, crucially, by conditioning the token probability of $\tau(e)$ on the reasoning $c$ and marginalizing over these critiques, the CoT verifier naturally generates a posterior predictive distribution $p_\theta(E_t = e \mid \textbf{x}_{\leq t}) = \textstyle\sum \limits_{c} p_\theta(e \mid \mathbf{x}_{\leq t}, c) \, p_\theta(c \mid \mathbf{x}_{\leq t})$ over the decisions $e$. This formulation treats different CoTs $c$ as distinct justifications generated by the judge-LM backing its final verification judgment about the current step $s_t$, ensuring that the judgment is supported by the reasoning. Unlike methods that generate explanations after a model reaches a conclusion \citep{zheng2023judging, wang2023pandalm}, our approach integrates reasoning into the decision-making process. Furthermore, our method leverages the full (approximated) posterior predictive distribution to obtain a probabilistic measure of confidence across different outcomes, differing from self-consistency \citep{wang2022self}, which relies solely on majority voting over samples from the distribution for point estimation.

\section{Uncertainty Quantification on Math Reasoning Verification}
\label{ref:experimental_setup}
In this section, we detail the experimental setup for measuring the performance of different uncertainty quantification methods $u$ on step-wise math reasoning verification.  
\subsection{Datasets}

We conduct experiments on the process supervision dataset PRM800K \citep{lightman2023let}, which consists of 800k step-level correctness labels for model-generated solutions to problems from the MATH \citep{hendrycks2021measuring}, a challenging competition math dataset. During annotation, data labelers were presented with step-by-step solutions to MATH problems pre-generated by GPT-4 and assigned each step a label of positive, negative, or neutral. We consider the negative class $y=1$,  indicating the presence of an error, while converting all other labels to $y=0$.
Due to computational constraints, we select a subset of 150 out of 500 questions from the test split, resulting in a total of 1,152 steps for judge-LM verification. Among these annotated steps, 129 (11.2\%) are labeled as containing errors ($y=1$), leading to a class imbalance. The average number of steps per solution is 7.7.

\subsection{Models}

We use \textsc{Qwen2-Math-72B-Instruct}\footnote{\url{https://qwenlm.github.io/blog/qwen2-math/}} as our judge-LM. Qwen2-Math-72B-Instruct is a specialized mathematical language model built upon the \textsc{Qwen2} LLMs, achieving strong performance on math benchmarks such as GSM8K \citep{cobbe2021training}. We use the instruct version, as empirically it demonstrates a better understanding of the task of verifying mathematical solutions. Following \citet{Farquhar2024se}, for each step $s_t$ we sample a single generation at low temperature (0.1) as the predicted label $\hat{e}_t$, which is used to assess the accuracy with the ground truth label $y_t$, and 10 generations at high temperature (1.0) producing diverse reasoning paths $c$ and decisions $e$ for clustering.

\subsection{Baseline Methods}
We compare CoT Entropy against the following baseline methods. \textbf{Naive Entropy} is the length-normalized average log token probabilities across the same number of generations as for CoT Entropy. \textbf{P(True)} \citep{kadavath2022language} estimates uncertainty by prompting an LM to compare a main answer with ‘brainstormed’ alternatives and using the predicted probability that the main answer is ‘True'. We use 0-shot prompting in our experiments where judge-LM self-reflects how likely its greedy verification result is compared to 10 alternatives. \textbf{SEU} \citep{grewal2024improving} 
is an embedding-based approach that measures uncertainty via the average pairwise cosine similarity of the embeddings of the generated responses. The intuition is that uncertain responses exhibit lower cosine similarity (greater semantic diversity) among outputs. This method relies on sentence embedding models to map semantically similar responses to nearby points in the embedding space. We use \textsc{MiniLM} \citep{wang2020minilm} and \textsc{NV-Embed} \citep{lee2024nv} as embedding models for SEU. 
\textbf{CoT Entropy (Discrete)} is a discrete variant of our proposed CoT Entropy. Following semantic entropy \citep{Farquhar2024se}, it replaces token probabilities in Eq.~\ref{eq:cot_entropy} with empirical decision class frequencies. As it does not rely on probabilities, it is a black-box method.
Additionally, we include a \textbf{Random} baseline, which samples an uncertainty score between 1 and 0 for each example. For all uncertainty methods, we run experiments with 5 random seeds and report the mean and standard deviation of their performance.

\subsection{Evaluation metrics}

We evaluate uncertainty quantification methods using the following metrics, each based on an uncertainty score estimated by $u_t$ assessed against reference labels $y_t$.  

\paragraph{AUROC.} Following previous works \citep{Farquhar2024se, kossen2024semantic}, we use the area under the receiver operating characteristic curve (AUROC) for the binary event that a given verification is correct. AUROC ranges from 0 to 1, where 1 indicates a perfect classifier and 0.5 corresponds to an uninformative classifier. \textbf{AUPRC.} We also include the area under the precision-recall curve (AUPRC), where the baseline for an uninformative classifier is equal to the proportion of positive samples in the dataset, which in our case, corresponds to the proportion of correctly verified steps.

\paragraph{AU-F1C and Rejection-F1.} We use selective performance scores to evaluate how well a model's uncertainty estimates can reject verification steps likely to lead to errors. While accuracy is commonly used in previous works \citep{Farquhar2024se}, we opt for F1-score due to the observed class imbalance (11.2\% positive labels) – ensuring a fairer evaluation – while also highlighting the importance of identifying positive cases, which indicate that an intermediate step contains an error. \textbf{Rejection-F1} measures the F1-score on the retained steps after filtering by uncertainty. We define “Rejection-F1 at \textit{X}\%” as the F1-score computed on the most confident \textit{X}\% of inputs, as determined by the uncertainty estimates. “Rejection-F1 at \textit{100}\%” yields the same performance across all methods, equivalent to an uninformative classifier. The area under the F1 curve (\textbf{AU-F1C}) serves as a summary statistic across multiple uncertainty thresholds, capturing the potential F1 improvement a user would experience when filtering out the most uncertain steps. In our experiments, AU-F1C is computed from 30\% onward to avoid low-data regime below this threshold, and Rejection-F1 is reported for thresholds from 60\% to 100\%.


\section{Evaluation Results}
We evaluate the different uncertainty quantification methods $u$ initialized with \textsc{Qwen-2-Math-72B-Instruct} on a subset of PRM800K test set as detailed in \S\ref{ref:experimental_setup}. We show the main evaluation results in \S\ref{sec:main_result} and analyze the sources of uncertainty in step-wise verification in \S\ref{sec:uncertainty_decomposition}.

\subsection{Does CoT Entropy inform confidence in verification?}
\label{sec:main_result}
We report AUROC, AUPRC, AU-F1C with standard deviation across the five runs in table~\ref{tab:metric_results}. Overall, CoT Entropy achieves the highest performance across the three metrics, indicating its effectiveness in distinguishing between correctly and incorrectly verified steps. We note that the baseline values for AUROC and AUPRC differ: AUROC is approximately 0.5 for an uninformative classifier that randomly assigns prediction correctness labels, while the AUPRC baseline corresponds to the proportion of correctly verified steps (0.786). For more details on verification performance see table~\ref{tab:verification_performance} in App.~\ref{app:verification}.

\begin{table}[ht]
\centering
\caption{Comparison of different uncertainty estimation methods.}
\label{tab:metric_results}
\scalebox{0.9}{
\begin{tabular}{l ccc}
\toprule
\textbf{Method} & \textbf{AUROC} & \textbf{AUPRC} & \textbf{AU-F1C} \\
\midrule
Random & $0.521_{±0.024}$ & $0.786_{±0.010}$ & $0.279_{±0.019}$\\
Naive Entropy   & $0.408_{±0.013}$ & $0.742_{±0.004}$ & $0.265_{±0.009}$\\
P(True) & $0.329_{±0.015}$ & $0.671_{±0.007}$ & $0.302_{±0.007}$\\
SEU-Minilm & $0.661_{±0.029}$ & $0.862_{±0.055}$ & $0.304_{±0.010}$\\
SEU-NV-Embed  & $0.616_{±0.011}$ & $0.759_{±0.009}$ & $0.330_{±0.010}$\\
CoT Entropy  & $\textbf{0.680}_{±0.017}$ & $\textbf{0.885}_{±0.005}$ & $\textbf{0.348}_{±0.009}$\\
\bottomrule
\end{tabular}}
\end{table}

\begin{figure}[h]
    \centering
    \includegraphics[width=0.6\textwidth]{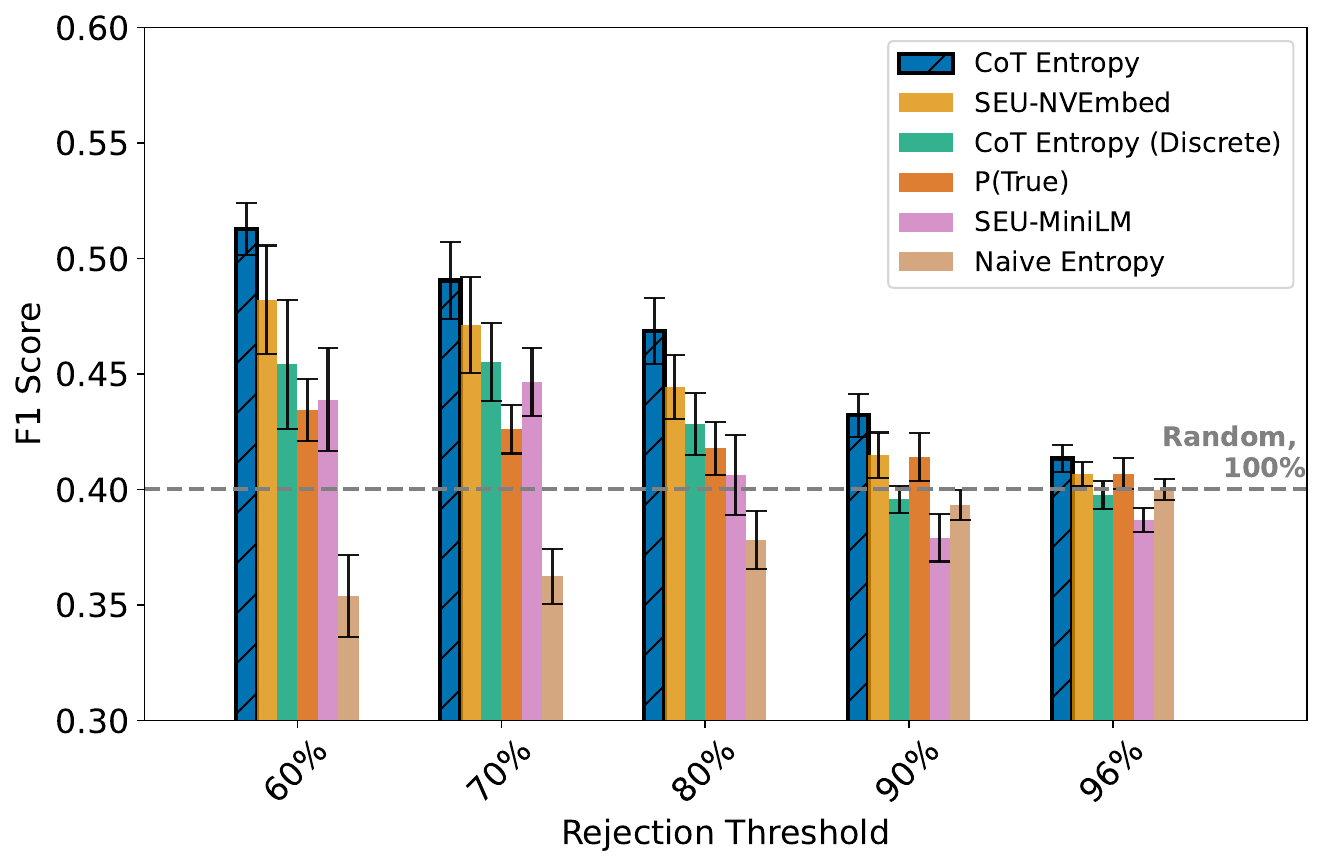}
    \caption{\textbf{Rejection-F1 for different uncertainty quantification (UQ) methods.} The bars represent the F1-Score on the retained examples, corresponding to the $\textit{X}\%$ most confident examples as determined by the UQ method, at the rejection threshold on the $\textit{x}$-axis. CoT Entropy outperforms leading baselines in detecting correctness of step-wise verification for intermediate reasoning traces for solving math problems. Results are averaged over five runs.} 
    \label{fig:uncertainty_comparison}
\end{figure}

Fig.~\ref{fig:uncertainty_comparison} presents the Rejection-F1 of various uncertainty quantification methods. The \textit{x}-axis represents the rejection threshold, ranging from 60\% to 96\%, while the \textit{y}-axis shows the F1-score on the retained examples, corresponding to the most confident \textit{X}\% of the total examples. The dashed line in the background represents the 100\% threshold (1,152 steps in total), where no uncertainty method is applied. This reflects the model’s performance on the original unfiltered test set, and also serves as a reference performance equivalent to the random baseline. We note that with 60\% threshold, the error bars are larger, due to the lower number of steps (691) in this bin.
If an uncertainty method is effective, predictions on the confident subset should achieve a higher performance score than those on the excluded subset, with Rejection-F1 improving as more inputs are rejected. This trend is evident in the plot, where Rejection-F1 increases from right to left, indicating that these methods generally produce meaningful confidence estimates for step-wise verification.

We observe that CoT Entropy achieves the highest Rejection-F1 among all thresholds, indicating the method's effectiveness in detecting verification errors. 
The naive entropy baseline, which lacks clustering, performs worse than random, highlighting the importance of structuring outputs into meaningful clusters for entropy-based methods. CoT Entropy (Discrete) underperforms CoT Entropy, suggesting that access to probabilities enhances entropy estimation. This differs with semantic entropy \citep{Farquhar2024se}, where little performance gap is observed between discrete and non-discrete variants. Aside from the fixed number of clusters in our setup, we hypothesize that the difference arises from the complexity of the math verification task. In our setting, probabilities capture nuances conditioned on preceding reasoning, meaning that even when two outputs reach the same decision, their probabilities may differ based on the reasoning path taken.

Comparing the embedding-based SEU methods, we observe that \textsc{NV-Embed} outperforms \textsc{MiniLM}, indicating that a higher-quality embedding model maps outputs to a more meaningful continuous space, leading to better estimation of semantic diversity. Notably, CoT Entropy outperforms SEU, which relies on an external embedding model. This suggests that when the judge-LM produces a meaningful predictive distribution, its token probabilities alone can enable more effective uncertainty quantification without relying on external models for heuristics.

Nevertheless, reward verification on reasoning is a challenging task, and the Rejection-F1 performance indicates room for improvement. However, we have shown that the proposed CoT Entropy method effectively informs the judge-LM’s confidence in step-wise verification. Using selective verification for steps with lower uncertainty improves verification reliability for reward models, particularly in complex reasoning tasks.

\subsection{How does different uncertainty types perform in the verification task?}
\label{sec:uncertainty_decomposition}

To understand the relevance of different sources of uncertainty present in the step-wise verification setup, we decompose the total predictive uncertainty associated with each data point into an epistemic and an aleatoric components \citep{malinin2020uncertainty}:
\begin{equation}
\begin{split}
\underbrace{\mathcal{I}\big[E_{t}, \bm{c} | Q, \mathcal{D}\big]}_{\text{Epistemic Uncertainty}} =&\ \underbrace{\mathcal{H}\big[{\tt P}(E_{t} |  Q, \mathcal{D})\big]}_{\text{Predictive Uncertainty}} - \underbrace{\mathbb{E}_{{\tt q}(c | \mathcal{D})}\big[\mathcal{H}[{\tt P}(E_{t} | Q; 
\bm{c})]\big]}_{\text{Aleatoric Uncertainty}}.
\end{split}
\label{eqn:mi}
\end{equation}



\begin{figure}
    \centering
 \includegraphics[width=0.6\linewidth]{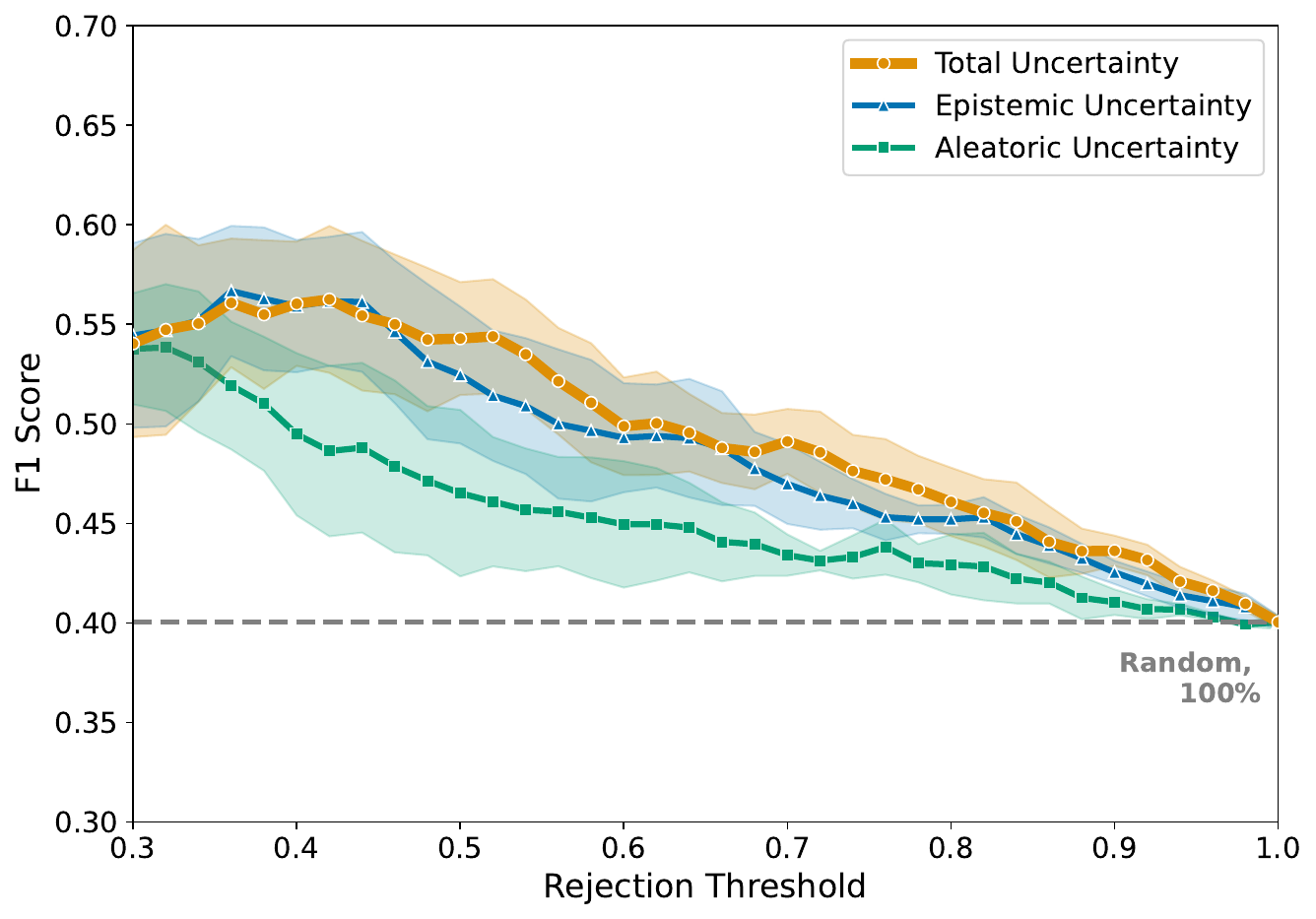}
    \caption{\textbf{Decomposition of the total predictive uncertainty}. As expected for a verification task, the total uncertainty better identifies verifier's mistakes. Nonetheless, epistemic uncertainty is almost as good, revealing that, for math reasoning verification, most identified errors are associated with model's knowledge uncertainty, as opposed to label noise. }
    \label{fig:uncertainty_breakdown}
\end{figure}

Equation \ref{eqn:mi} computes the Epistemic Uncertainty of a Bayesian model as the Mutual Information between the predicted variable and the hypotheses variable \citep{10.1214/aoms/1177728069}. In our case, these variables are, respectively, the step-wise verification $E_{t}$ and the rationales $c$. Furthermore, $q(c | \mathcal{D})$ represents the posterior over rationales, which is given by our generator model conditioned on the in-context dataset $\mathcal{D}$. Empirically, we approximate this posterior over rationales with Monte-Carlo sampling, which allows us to easily approximate the posterior predictive distribution and the terms in Equation \ref{eqn:mi}.


We employ the three different types of uncertainty in the verification task and report the Rejection-F1 on Figure \ref{fig:uncertainty_breakdown}. First, we highlight that all uncertainty sources improve over a random baseline, showing that the uncertainty estimations over the rationales are indeed effective in identifying the questions where the verifier potentially does not know the answer. We also observe that the predictive uncertainty presents the best Rejection-F1. This is expected given the nature of a verification task: we hope to find \textit{any} point that may lead to a prediction error, regardless of the underlying reason. This contrasts with other problem settings, such as Bayesian Active Learning, where the focus is on epistemic uncertainty as a way to improve model's knowledge \citep{10.1214/aoms/1177728069, 10.5555/3305381.3305504, NEURIPS2024_d5e256c9}.

Furthermore, we observe that the epistemic uncertainty (representing the model's uncertainty, or lack of knowledge) better correlates with the prediction errors than the aleatoric uncertainty (representing the uncertainty about the label-generating process -- for instance, a disagreement between the labelers). In fact, epistemic uncertainty is almost as good as the total uncertainty in predicting the verifier prediction errors. Overall, this suggests an interesting insight about the problem setting: most of the identified errors are associated with model's uncertainty -- indeed, improving the capacity of LLMs for math reasoning remains an active area of research.

\section{Discussion}
In this work, we explored using uncertainty quantification (UQ) as a principled way to improve the reliability of step-wise verification using generative reward models on math reasoning tasks. We extend the application of UQ beyond standard QA to the domain of complex reasoning. We proposed CoT Entropy, a novel method that informs the judge-LM’s confidence in verifying step-wise reasoning process in a PRM dataset, outperforming other baseline UQ methods. One limitation of our work is that the prompt for judge-LM can be further optimized. For future work, we aim to leverage the estimated uncertainty to address and potentially mitigate reward hacking during RL training while also aiding inference-time search. Additionally, we plan to fine-tune a generative RM to assess whether it can produce more accurate uncertainty estimates for verification.



\bibliography{neurips_2024}
\bibliographystyle{plainnat}


\appendix
\section{Appendix}
\subsection{Prompt Template}
\label{app:prompt}

We show the template we use to prompt LLMs for step-wise verification in Fig.~\ref{fig:verification_template}. During our experiments, we have tested various prompting templates and observed that it is crucial to use a prompt that increases the JSON parsing success rate, as it serves an indicator of the model's understanding of the verification task.

\begin{figure*}[!th]
\begin{tcolorbox}[
    colback=white, 
    colframe=black, 
    title=\textbf{Template for Step-wise Verification}, 
    fonttitle=\bfseries\large, 
    arc=4mm, 
]
You are a professional mathematician. Given a problem and the previously proposed steps, your task is to evaluate whether the next proposed step contains any errors in relation to the preceding steps, giving your reasoning. If there is an error, state ``yes". If there is no error, state ``no". Focus solely on the transition from the previous step to the next.

The evaluation should adhere to the following criteria: \\

1. Accuracy: Verify all calculations, including algebraic manipulations and numerical computations, are correct.

2. Logical Progression: Ensure the next proposed step follows logically from the previous step, applying mathematical rules, theorems, or formulas correctly, and making reasonable observations. Note: Omit this criterion if the step being evaluated is the first step, as there are no preceding steps to compare.

3. Step-by-Step Focus: Evaluate only the immediate transition from the previous step to the next proposed step. Do not mark the next step as incorrect for not performing an action that should logically occur in a future step. \\ \\ 
Problem: \{user\_problem\} \\
Preceding Steps: \{solution\_so\_far\} \\ \\ 
Next Proposed Step to be Evaluated: \{next\_step\} \\ \\
Now, generate your response in the following JSON format. Give your reasoning in short and concise sentences after ``reasoning”, then output your final evaluation after ``has\_error”.\\ \\
\{``reasoning": "Your reasoning here.", ``has\_error": "yes/no"\}

\end{tcolorbox}
\end{figure*}\label{fig:verification_template}

\subsection{Verification Performance}
\label{app:verification}
\begin{table}[ht]
\centering
\caption{Comparison of verification performance with Qwen2-Math-72B-Instruct, prompted with or without CoT. Although No-CoT achieves higher accuracy, this is largely due to the model predominantly predicting a label of 0 (no error). Since errors constitute the minority class in this imbalanced dataset, F1 score provides a more meaningful evaluation, as it better reflects the model’s ability to detect errors.}
\label{tab:verification_performance}
\scalebox{0.9}{
\begin{tabular}{l cc}
\toprule
\textbf{Method} & \textbf{F1} & \textbf{Acc.} \\
\midrule
No-CoT prompted & $0.352_{±0.010}$ & $0.838_{±0.003}$ \\
CoT prompted & $0.400_{±0.003}$ & $0.774_{±0.002}$ \\
\midrule
Predicting all 0s & $0.0 $ & $0.889$ \\
Predicting all 1s & $0.197 $ & $0.112$ \\
\bottomrule
\end{tabular}}
\end{table}
 
\end{document}